\documentclass[12pt]{article}

\usepackage{amssymb,amsmath,amsfonts,latexsym,graphicx}		
\usepackage[linesnumbered,ruled,vlined]{algorithm2e}
\usepackage[hidelinks]{hyperref}
\usepackage{cleveref}
\usepackage[numbers]{natbib}
\usepackage{array}
\usepackage{subcaption}
\usepackage{mathrsfs}  
\usepackage{graphicx} 
\usepackage{tabularx} 
\usepackage{caption} 
\usepackage{float}
\usepackage{wrapfig}
\usepackage{placeins}

\crefname{figure}{Figure}{Figures}
\crefname{equation}{Eq.}{Eqs.}
\crefname{table}{Table}{Tables}
\crefname{algorithm}{Algorithm}{Algorithms}
\crefname{section}{Section}{Sections}

\newcounter{daggerfootnote}
\newcommand*{\daggerfootnote}[1]{%
    \setcounter{daggerfootnote}{\value{footnote}}%
    \renewcommand*{\thefootnote}{\fnsymbol{footnote}}%
    \footnote[2]{#1}%
    \setcounter{footnote}{\value{daggerfootnote}}%
    \renewcommand*{\thefootnote}{\arabic{footnote}}%
    }

\begin{document}

\begin{center}
\Large \bf CAVE-Net: Classifying Abnormalities in Video Capsule Endoscopy \rm

\vspace{1cm}

\def\thefootnote{* }\footnotetext{Equal Contribution}

\large Ishita~Harish \textsuperscript{1,*}, Saurav~Mishra \textsuperscript{1,*}, Neha~Bhadoria \textsuperscript{1}, Rithik~Kumar \textsuperscript{2}, Madhav~Arora \textsuperscript{1}, Syed~Rameem~Zahra \textsuperscript{3}, Ankur~Gupta \textsuperscript{1,}\daggerfootnote{Ankur Gupta is the corresponding author.}

\vspace{0.5cm}
\normalsize
\textsuperscript{1} Netaji Subhas University of Technology, New Delhi \\
\textsuperscript{2} Helios Tech Solutions Pvt Ltd, Gurugram, Haryana \\
\textsuperscript{3} Centre for Artificial Intelligence and Machine Learning, SKUAST-K, J\&K, India

\vspace{5mm}

Email: {
\tt{harishishita@gmail.com},
\tt{sauravmishra2612@gmail.com},
\tt{nehabhadoria294@gmail.com},
\tt{shoraj617@gmail.com},
\tt{eragon.cube@outlook.com},
\tt{syed.zahra@skuastkashmir.ac.in},
\tt{agupta4@cs.iitr.ac.in}}

\vspace{1cm}

\end{center}

\abstract{Accurate classification of medical images is critical for detecting abnormalities in the gastrointestinal tract, a domain where misclassification can significantly impact patient outcomes.
We propose an ensemble-based approach to improve diagnostic accuracy in analyzing complex image datasets.
Using a Convolutional Block Attention Module along with a Deep Neural Network, we leverage the unique feature extraction capabilities of each model to enhance the overall accuracy.
The classification models, such as Random Forest, XGBoost, Support Vector Machine and K-Nearest Neighbors are introduced to further diversify the predictive power of proposed ensemble.
By using these methods, the proposed framework, CAVE-Net, provides robust feature discrimination and improved classification results.
Experimental evaluations demonstrate that the CAVE-Net achieves high accuracy and robustness across challenging and imbalanced classes, showing significant promise for broader applications in computer vision tasks.}

\vspace{7mm}


\section{Introduction}\label{sec1}

Capsule endoscopy~\cite{hale2014capsule} is a minimally invasive procedure that involves swallowing a small, pill-sized camera that captures images of the gastrointestinal tract, enabling doctors to examine areas that are difficult to reach with traditional endoscopy methods.

This paper proposes an ensemble approach that integrates multiple techniques of machine learning and deep learning for the purpose of robust feature extraction and optimized decision-making in classification tasks.
We make use of the datasets~\cite{Handa2024Training,Handa2024} provided by the Capsule Vision 2024 Challenge~\cite{handa2024capsulevision2024challenge}, which comes with a rich diversity of medical images, to test our proposed models for automatic classification of abnormalities in Video Capsule Endoscopy frames.
The main goal of this research is to utilize the feature-selective strengths of well-established classifiers to make final classification decisions with high accuracy.

We evaluate the performance of our ensemble method in dealing with complex datasets with imbalanced class distributions through a systematic series of experiments, thus demonstrating its potential for use in a wide range of challenging scenarios.

\section{Methods}\label{sec2}

To address class imbalance, image augmentation is used to create a balanced training dataset of approximately $96,000$ images across ten classes.
A pre-trained ResNet50~\cite{he2015deepresiduallearningimage} model is adapted as an autoencoder~\cite{bank2021autoencoders} for feature extraction, generating latent space representations.
A decoder is used to reconstruct images from this latent space to validate the extracted features by comparing them with the original images. 
A small fraction of these reconstructed images is added to the training dataset to increase population diversity.

Latent representations are classified using a deep neural network (DNN) and Syn-XRF, an ensemble of SVM~\cite{708428}, Random Forest~\cite{Breiman2001}, KNN~\cite{Peterson:2009}, and XGBoost~\cite{Chen_2016}, employing majority voting.
Additionally, a ResNet based model, integrated with a Convolutional Block Attention Module (CBAM)~\cite{woo2018cbamconvolutionalblockattention}, is applied in parallel on the training dataset to refine feature representations and enhance abnormality detection.
The proposed framework, CAVE-Net, combines DNN, Syn-XRF, and CBAM-enhanced ResNet into a robust ensemble, addressing dataset challenges and delivering precise classifications.

\subsection{Dataset Specification}\label{dataset-spec}
The Capsule Vision 2024 Challenge~\cite{handa2024capsulevision2024challenge} dataset is organised with three publicly available VCE datasets: SEE-AI~\cite{yokote2024small}, KID~\cite{koulaouzidis2017kid}, and Kvasir-Capsule~\cite{smedsrud2021kvasir}, and a private dataset from AIIMS~\cite{goel2022dilated}.

The training and validation datasets~\cite{Handa2024Training} contained $37,607$ and $16,132$ frames, respectively, categorized into ten classes\textemdash \textit{Angioectasia, Bleeding, Erosion, Erythema, Foreign Body, Lymphangiectasia, Polyp, Ulcer, Worms} and \textit{Normal}. The images were preprocessed by cropping, resizing to $224\times224$ pixels, and ensuring consistency in their labels.

\begin{figure}[htp]
    \centering 
    \caption{Representative images from each class in the dataset.}
    
    \begin{minipage}{0.1666\textwidth}
        \centering
        \includegraphics[width=\textwidth]{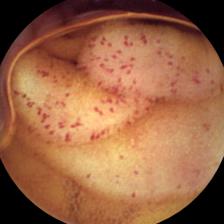}
        \subcaption{Angioectasia}
    \end{minipage} \hspace*{0.05em}
    \begin{minipage}{0.1666\textwidth}
        \centering
        \includegraphics[width=\textwidth]{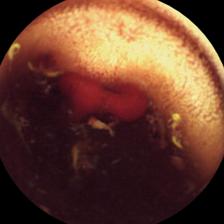}
        \subcaption{Bleeding}
    \end{minipage} \hspace*{0.05em}
    \begin{minipage}{0.1666\textwidth}
        \centering
        \includegraphics[width=\textwidth]{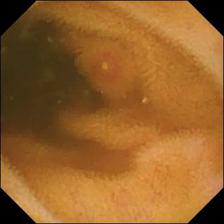}
        \subcaption{Erosion}
    \end{minipage} \hspace*{0.05em}
    \begin{minipage}{0.1666\textwidth}
        \centering
        \includegraphics[width=\textwidth]{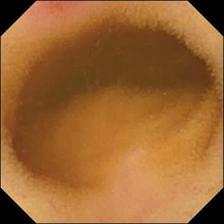}
        \subcaption{Erythema}
    \end{minipage} \hspace*{0.05em}
    \begin{minipage}{0.1666\textwidth}
        \centering
        \includegraphics[width=\textwidth]{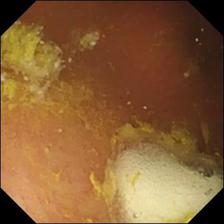}
        \subcaption{Foreign Body}
    \end{minipage} \\[1ex]
    \begin{minipage}{0.1666\textwidth}
        \centering
        \includegraphics[width=\textwidth]{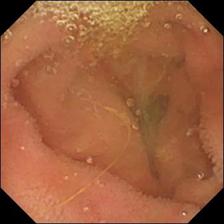}
        \subcaption{Lymphangiectasia}
    \end{minipage} \hspace*{0.05em}
    \begin{minipage}{0.1666\textwidth}
        \centering
        \includegraphics[width=\textwidth]{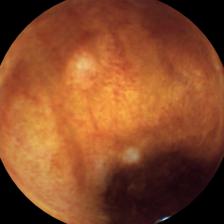}
        \subcaption{Polyp}
    \end{minipage} \hspace*{0.05em}
    \begin{minipage}{0.1666\textwidth}
        \centering
        \includegraphics[width=\textwidth]{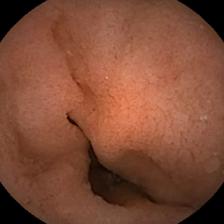}
        \subcaption{Ulcer}
    \end{minipage} \hspace*{0.05em}
    \begin{minipage}{0.1666\textwidth}
        \centering
        \includegraphics[width=\textwidth]{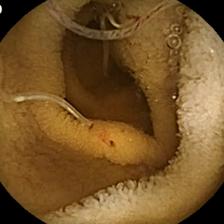}
        \subcaption{Worm}
    \end{minipage} \hspace*{0.05em}
    \begin{minipage}{0.1666\textwidth}
        \centering
        \includegraphics[width=\textwidth]{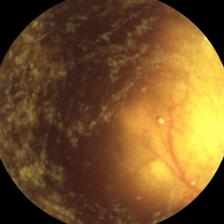}
        \subcaption{Normal}
    \end{minipage}
    
\label{fig:class-rep}
\end{figure}

The test dataset~\cite{Handa2024} comprised $4,385$ medically annotated VCE frames from All
India Institute of Medical Sciences (AIIMS) Delhi, derived from over 70 patient videos.
This dataset, de-identified and ethically approved by the Department of Gastroenterology and HNU, AIIMS Delhi, focused on nine abnormalities (excluding the normal class).
A representative image from each class is shown in~\cref{fig:class-rep}.

\subsection{Data Augmentation}\label{data-aug}
To address the class imbalance in the training dataset, where the \textit{Normal} class dominated and rarer classes like \textit{Worms} and \textit{Ulcer} had minimal samples, data augmentation was employed.
The goal was to ensure a minimum of $7,500$ samples per class, ensuring that each abnormality is well represented in training.

The augmentation set $\mathcal{O}$ includes the following transformations:$$\mathcal{O} = \{\mathcal{F},~\mathcal{R},~\mathcal{Z},~\mathcal{S},~\mathcal{B},~\mathcal{N},~\mathcal{C}\}$$where $\mathcal{F}$ randomly applies horizontal and vertical flips to the image to account for orientation invariance in gastrointestinal imaging.
$\mathcal{R}$ introduces rotations within $\pm 20^\circ$, mimicking different viewpoints of the capsule as it navigates the tract.
$\mathcal{Z}$ represents zoom transformations, both zoom-in and zoom-out, with scaling factors of $\pm 15\%$, providing a varied perspective by focusing closely on abnormalities or offering a wider context.
The transformation $\mathcal{S}$ applies shearing with angles of up to $\pm 10^\circ$, simulating slight distortions caused by natural movement and curvature of the gastrointestinal tract.
$\mathcal{B}$ represents Gaussian blur, applied using a kernel size of 5 and $\sigma \in [0.1, 1.0]$, to mimic the blurring effects commonly seen due to motion artifacts.
Similarly, $\mathcal{N}$ adds Gaussian noise with a standard deviation of $0.05$, emulating sensor-induced noise, thereby enhancing the model’s robustness to pixel-level variations.
Finally, $\mathcal{C}$ performs random cropping between $85\%$\textendash$100\%$ of the original image, followed by resizing to $224 \times 224$ pixels, introducing spatial variability while maintaining the structural integrity of the abnormalities.
These transformations were applied in random combinations of two to four operations per image to ensure diversity and realism in the augmented dataset, effectively addressing class imbalance while preserving crucial anatomical features. 
    
    
    
The distribution of the training and validation datasets before and after augmentation are shown in~\cref{tab:table1}.

\begin{table*}[t]
\centering
 \caption{Dataset’s original and post-augmentation class-wise instance counts.}

\label{tab:table1}
\resizebox{0.85\textwidth}{!}{
\begin{tabular}{cccccc}
\hline
\textbf{Class}              & \textbf{\begin{tabular}[c]{@{}c@{}}Training \\ Examples\end{tabular}} & \textbf{\begin{tabular}[c]{@{}c@{}}Validation \\ Examples\end{tabular}} & \textbf{\begin{tabular}[c]{@{}c@{}}Training Examples \\ (After Augmentation)\end{tabular}} \\ \hline
\textbf{Angioectasia}       & 1,154  & 497   & 7,500 \\ 
\textbf{Bleeding}           & 834    & 359   & 7,500 \\ 
\textbf{Erosion}            & 2,694  & 1,155 & 7,500 \\ 
\textbf{Erythema}           & 691    & 297   & 7,500 \\ 
\textbf{Foreign Body}       & 792    & 340   & 7,500 \\ 
\textbf{Lymphangiectasia}   & 796    & 343   & 7,500 \\ 
\textbf{Normal}             & 28,663 & 12,287 & 28,663 \\ 
\textbf{Polyp}              & 1,162  & 500   & 7,500 \\ 
\textbf{Ulcer}              & 663    & 286   & 7,500 \\ 
\textbf{Worm}               & 158    & 68    & 7,500 \\ \hline
\textbf{Total}              & \textbf{37,607} & \textbf{16,132} & \textbf{96,163} \\ \hline
\end{tabular}}
\end{table*}


\subsection{Autoencoder with ResNet}\label{subsec2}

An autoencoder architecture based on ResNet50~\cite{he2015deepresiduallearningimage} was developed to extract feature representations from images and to generate reconstructed images.
The autoencoder consists of two components, an encoder and a decoder.
The encoder utilizes a ResNet50 backbone, pre-trained on {ImageNet~\cite{5206848}} dataset.
The architecture includes 50 layers, comprising convolutional layers, batch normalization layers, ReLU activations and skip connections, allowing the model to learn identity mappings.

The ResNet encoder processes the input images \( I \) and generates a latent representation \( z \) as shown in~\cref{eq:latent}. The fully connected layer of ResNet was replaced with a linear layer, which outputs a latent space representation of size 1024.
\begin{equation}
    z = Encoder(I) \in \mathbb{R}^{1024} \label{eq:latent}
\end{equation}

The decoder takes the latent representation \( z \) and reconstructs the original image \( \hat{I} \). The decoding process, as shown in \cref{eq:decoder}, comprises of several transposed convolutional layers, which progressively upscale the latent representation back to the original image dimensions.
\begin{equation}
    \hat{I} = Decoder(z) \label{eq:decoder}
\end{equation}
The reconstruction is evaluated using Mean Squared Error (MSE) loss, as defined in \cref{eq:mse_loss}, where \( N \) represents the number of pixels in the image ($224\times224\times3=150528$).
The model was trained over a maximum of 40 epochs, adjusting the weights via the Adam optimizer to minimize the reconstruction loss, and applying early stopping regularization.
\begin{equation}
    \mathcal{L} = \frac{1}{N} \sum_{i=1}^{N} (I_i - \hat{I}_i)^2 \label{eq:mse_loss}
\end{equation}

Upon completion of training, a portion of the reconstructed images generated by the autoencoder was merged back into the training dataset.
This integration aims to enhance the diversity of the training data, effectively increasing the dataset size and aiding subsequent classification tasks.
The latent space representations \( z \) generated from the validation set during the encoding process were stored and utilized for training models in the subsequent stages of the framework, CAVE-Net.
The final latent space representations contained condensed, relevant information about the input images, successfully reducing feature size from $150528$ to $1024$.

\subsection{Deep Neural Network}\label{subsec5}
The Deep Neural Network (DNN)~\cite{7867471} was designed to classify features extracted from the pre-trained autoencoder described in~\cref{subsec2}.
The architecture comprises multiple fully connected layers, utilizing nonlinear transformations and dropout regularization to transform the latent representations into class probabilities.
This design enables the model to effectively distinguish between complex image classes.

The DNN layers apply sequential linear transformations followed by ReLU activations. Dropout regularization is introduced after specific layers to prevent overfitting by randomly deactivating neurons during training. The output layer performs a linear transformation to compute class scores, which are converted into probabilities using the softmax function as shown in~\cref{eq:softmax}. The network minimizes the cross-entropy loss defined in~\cref{eq:cross_entropy_loss}, ensuring accurate classification by penalizing incorrect predictions.

\begin{equation} P(y_i \mid x) = \frac{e^{y_i}}{\sum_{j=1}^C e^{y_j}} \label{eq:softmax} \end{equation}

\begin{equation} \mathcal{L} = - \frac{1}{N} \sum_{i=1}^{N} \sum_{c=1}^{C} y_{ic} \log(P(y_c \mid x_i)) \label{eq:cross_entropy_loss} \end{equation}

Here, \(C\) is the number of classes, N is the number of training samples, and \(y_{ic}\) indicates the true class label.

The model was trained over 50 epochs with a batch size of 32, using the Adam optimizer with a learning rate of 0.001. A 5-fold cross-validation strategy was adopted to iteratively update model weights and evaluate performance. Validation accuracy was tracked throughout to assess generalization on unseen data.

The DNN can appropriately learn the complex feature representations from the high-dimensional latent space.
DNNs can model non linear decision boundaries, making them effective for distinguishing between classes that are not linearly separable.
By utilizing latent space features extracted from the autoencoder, the DNN refines these high-level representations for improved classification accuracy.

\subsection{Syn-XRF Model}\label{subsec3}
An ensemble model is implemented which consists of four distinct classifiers, SVM, RF, KNN, and XGBoost.
Each classifier generates probabilistic outputs for class labels based on their respective learning mechanisms.

The SVM is utilized to find the optimal hyperplane that separates the classes in the latent space representation.
The decision function of SVM can be mathematically expressed as shown in~\cref{eq:svm_decision}.
\begin{equation}
    f(x) = w^T \cdot X + b \label{eq:svm_decision}
\end{equation}
where \( w \), \( X \), and \( b \) represent the weight vector, input feature vector, and the bias term, respectively.
The optimization goal is to maximize the margin between the classes, as depicted in \cref{eq:svm_margin}.
\begin{equation}
    \texttt{maximize} \left(\frac{2}{||w||}\right)\quad\vert\quad y_i(w^T X_i + b) \geq 1 \quad \forall i
    \label{eq:svm_margin}
\end{equation}

The RF constructs multiple decision trees during training and outputs the mode of the classes (classification) or mean prediction (regression) of the individual trees.
The model was implemented with 100 decision trees, chosen to balance computational efficiency and model performance.
This configuration allows the ensemble to aggregate predictions from a diverse set of trees, reducing the risk of overfitting while maintaining high predictive accuracy.
The final class prediction can be represented as in \cref{eq:rf_prediction}.
\begin{equation}
    \hat{y} = \text{mode}(h_1(x), h_2(x), \dots, h_n(x)) \label{eq:rf_prediction}
\end{equation}
where \( h_i(\cdot) \) represents the decision of the \( i \)-th decision tree.

The KNN classifies instances based on the majority label of their \( k \) nearest neighbors in the feature space.
For our implementation, \(k=7\) was selected as the optimal value after experimenting with various k-values to balance model performance and computational efficiency.
This parameter ensures effective neighborhood aggregation while capturing local structures in the data.
The predicted label \( \hat{y} \) for a sample \( x \) can be represented as in \cref{eq:knn_prediction}.
\begin{equation}
    \hat{y} = \text{argmax}_j \sum_{i=1}^{k} I(y_i = j) \label{eq:knn_prediction}
\end{equation}
where \( I \) is the indicator function that returns one if the class label \( y_i \) equals \( j \) and zero otherwise.

The XGBoost applies gradient boosting techniques to enhance classification precision.
The model combines multiple weak learners (typically decision trees) to form a strong learner.
The final prediction for a sample can be expressed as in \cref{eq:xgb_prediction}.
\begin{equation}
    \hat{y} = \sum_{i=1}^{N} \alpha_i h_i(x) \label{eq:xgb_prediction}
\end{equation}
where \( h_i(x) \) represents the output of the \( i \)-th tree and \( \alpha_i \) is the weight associated with the tree, $i$.

The ensemble model aggregates the predictions from the aforementioned classifiers using a voting mechanism.
In our implementation, we utilized soft voting, which considers the predicted probabilities of each class for each classifier.
The final prediction can be mathematically described as shown in \cref{eq:ensemble_soft_vote}.
\begin{equation}
    \hat{y} = \text{argmax}_j \sum_{i=1}^{M} P(h_i(x) = j), \quad \text{for } j \in \text{Classes} \label{eq:ensemble_soft_vote}
\end{equation}
In this approach, each classifier contributes its unique strengths.
The RF model excels in handling high-dimensional data and mitigating overfitting by aggregating the results of multiple decision trees, providing robustness against noise.
XGBoost leverages gradient boosting, allowing for efficient handling of missing values and the incorporation of complex patterns through its tree-boosting framework, thus enhancing predictive accuracy.
The SVM is adept at finding optimal hyperplanes in high-dimensional spaces, making it effective for margin-based classification, particularly in cases with clear class separations.
Lastly, KNN offers simplicity and adaptability, providing intuitive classifications based on proximity in feature space, which is particularly useful for capturing local structures.
By combining these classifiers, we aim to harness their diverse strengths, leading to improved overall performance in classification tasks.
The combination of diverse classifiers within the ensemble effectively enhances the model's predictive performance on the validation set.


\subsection{CBAM-Enhanced ResNet Architecture}\label{subsec4}
We have implemented the \textit{Convolutional Block Attention Module} (CBAM) to enhance the feature extraction capabilities of our classification model.
CBAM is designed to focus on significant features by applying attention mechanisms across both spatial and channel dimensions, thereby improving the model's performance in distinguishing between different classes.
The architecture of the CBAM consists of two primary components- the Spatial Attention Module and the Channel Attention Module.

The \textbf{Spatial Attention Module (SAM)} focuses on identifying the most important spatial regions in the input feature map. Given an input feature map \( F \in \mathbb{R}^{C \times H \times W} \), where \( C \), \( H \), and \( W \) represent the number of channels, height, and width, respectively, SAM generates a spatial attention map \( M_s \). This is computed in~\cref{eq:spatial_attention}

\begin{equation}
    M_s = \sigma\left(\text{Conv}\left(\text{concat}\left(\text{max}(F), \text{avg}(F)\right)\right)\right) \cdot F \label{eq:spatial_attention}
\end{equation}

Here, \( \sigma \) denotes the sigmoid activation function, \( \text{Conv} \) represents a convolutional operation, and \( \text{max} \) and \( \text{avg} \) are the maximum and average pooling operations applied along the channel dimension. The concatenated feature maps are passed through a convolutional layer to generate the final attention map, which is then element-wise multiplied with the input feature map \( F \).

The \textbf{Channel Attention Module (CAM)} emphasizes the significance of different channels in the feature map by learning a channel attention vector. For an input feature map \( F \), the channel attention map \( M_c \) is calculated by~\cref{eq:channel_attention}

\begin{equation}
    M_c = \sigma\left(W_2 \cdot \text{ReLU}\left(W_1 \cdot \text{GlobalAvgPool}(F)\right)\right) \label{eq:channel_attention}
\end{equation}

In this equation, \( W_1 \) and \( W_2 \) are weight matrices corresponding to learnable parameters in the linear layers, and \( \text{GlobalAvgPool} \) denotes the global average pooling operation applied to \( F \). The resulting channel attention map \( M_c \) is then multiplied element-wise with \( F \) to enhance its channel-wise discriminative features.

\begin{equation}
    F' = M_c \cdot (M_s \cdot F) \label{eq:combined_attention}
\end{equation}
where \( F' \) represents the refined feature map after applying both attention mechanisms.
This refined feature map is then fed into the subsequent layers of the model for classification.

The backbone of the CBAM-Enhanced ResNet model is ResNet-18, which consists of several convolutional layers and residual blocks designed to capture hierarchical features effectively.
The CBAM is applied to the output of the final feature extraction layer of the ResNet.
The attention mechanism focuses on enhancing critical feature maps before they are passed to the fully connected layer.
A linear layer follows the CBAM to perform classification based on the refined feature maps, which gives the final predictions across classes.

CBAM also contributes to model interpretability, making it a valuable tool for applications where understanding the model's decision-making process is crucial. This interpretability is especially beneficial in fields like medical imaging, where small variations in image data can have significant implications for diagnosis. 

This model is employed for its ability to enhance feature representation through spatial and channel attention mechanisms.
This dual attention allows the model to focus on relevant features while minimizing the impact of less informative ones, which is crucial for image classification tasks.
By improving sensitivity to salient features, CBAM enhances classification accuracy and model interpretability, making it particularly beneficial in complex medical imaging scenarios where subtle distinctions between classes are critical for accurate diagnosis.

\subsection{CAVE-Net}\label{subsec6}
\begin{figure}[H]
     \centering
     \caption{Architecture of the proposed framework, CAVE-Net.}

    \hspace{1.2em}

    \includegraphics[width=0.9\linewidth]{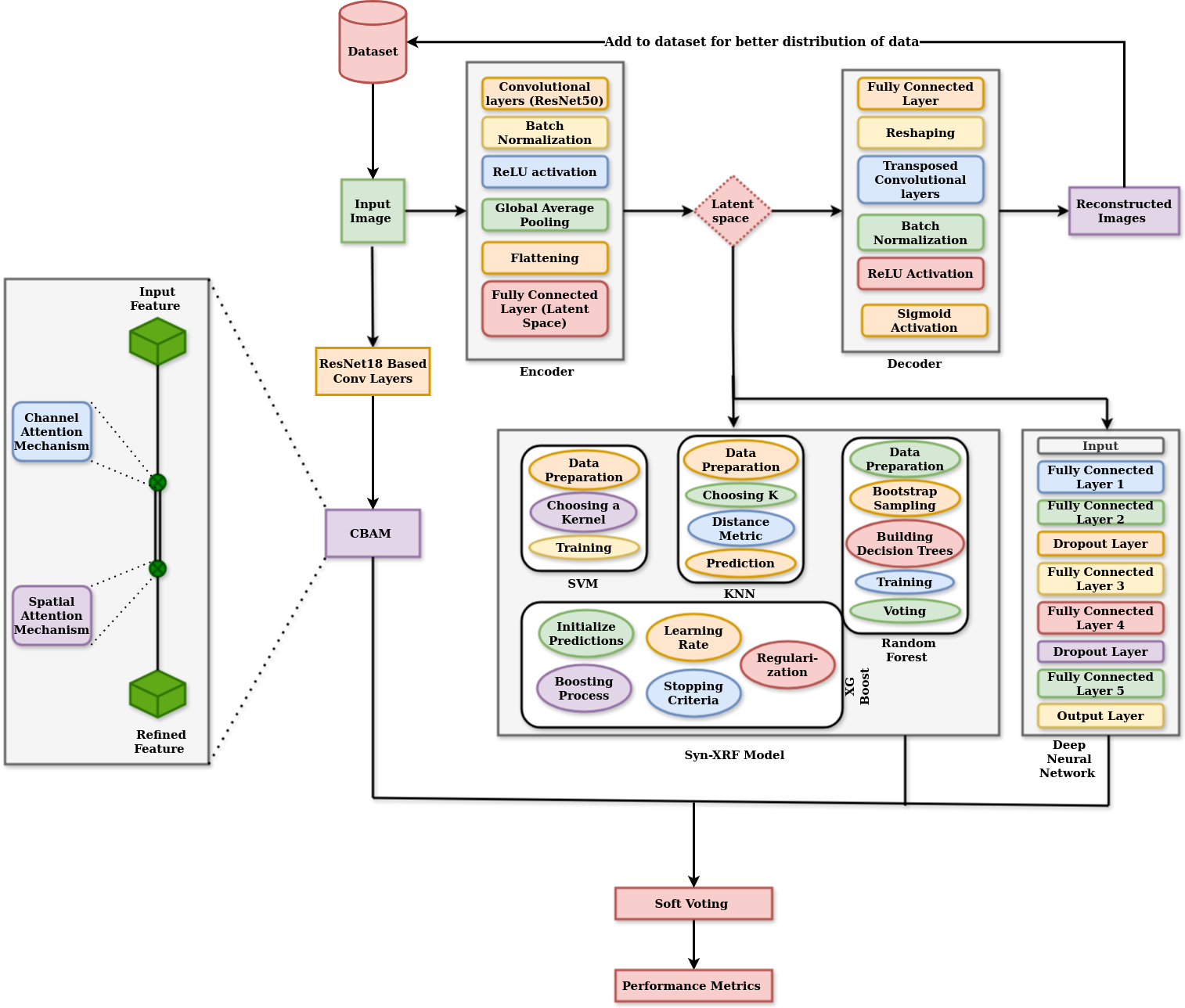}
    \label{fig:architecture}
\end{figure}

The final classification is determined using a soft voting mechanism~\cite{CAO2015275}, where the predicted probabilities out of the three separate models, CBAM, DNN, and Syn-XRF, contributed to the final decision. This classification model, as depicted in~\cref{fig:architecture}, has been named CAVE-Net. Each model is executed in parallel, enabling simultaneous computation of their respective output probabilities.

The CBAM is employed for its ability to enhance feature representation by focusing on both spatial and channel dimensions, thereby improving the model’s sensitivity to important features in the images.
The DNN is used because of to its capacity to learn complex non linear relationships within the extracted latent features, enabling effective classification.
The Syn-XRF is incorporated to leverage a different learning paradigm, thereby broadening the ensemble's ability to capture diverse patterns in the data.

By combining the outputs of these models through soft voting, we aimed to harness their individual strengths and mitigate their respective weaknesses.

This synergistic approach allows the ensemble to benefit from the nuanced feature extraction capabilities of CBAM, the deep learning power of DNN, and the complementary insights provided by the Syn-XRF.
The soft voting strategy aggregates the predicted probabilities for each class, selecting the class with the highest cumulative probability as the final output.

This method not only improves classification accuracy, but also increases robustness against overfitting and biases inherent to individual models, resulting in improved overall performance in the complex task of image classification.
The parallel execution of the three models ensures that the computational efficiency is maintained while taking full advantage of each model’s individual capabilities. This allows for faster processing and better scalability, which is crucial in real-world applications where large datasets are involved.

The combination of these diverse models, each focused on different aspects of the data, enables the system to adapt to variations in the dataset. The CBAM focuses on improving feature representation, the DNN learns complex patterns from the latent features, and the Syn-XRF brings a unique learning approach to the ensemble. By incorporating multiple paradigms, CAVE-Net is better equipped to handle the complexity and diversity of image data, leading to a more comprehensive classification system.

 The soft voting mechanism helps to create a robust classification model that balances the individual strengths of each approach, leading to a final output that is more accurate and reliable. By mitigating the potential weaknesses of each individual model, this approach ensures that CAVE-Net provides high-quality classification results even in challenging image-based tasks.

\section{Results}\label{sec3}

\subsection{Experimental Setup } 
The model  has been trained and tested with Python 3.10 on Ubuntu 20.04.6 LTS, using 256 GB of memory, an Intel Xeon Silver 4210R CPU (40 cores) and a single NVIDIA RTX A5000 GPU.
The model was trained on augmented images and evaluated on Capsule Vision 2024 Challenge Validation Dataset.

\subsection{Achieved results on the dataset} 
\label{subsec1}

CAVE-Net's performance was evaluated on both training and validation datasets, and it achieved impressive results in each case. For the training data, the framework attained an accuracy of 91\%, significantly exceeding the individual metrics obtained by the constituent models, as represented in Table \ref{tab:results1}. Similarly, for the validation dataset, we observed the overall accuracy of the CAVE-Net architecture to be 89.93\%, outperforming the constituent frameworks, as depicted in Table \ref{tab:results2}.

\begin{table*}[htbp]
\centering
\caption{Comparison of CAVE-Net Performance Metrics in the Training Dataset.}
\label{tab:results1}
\resizebox{0.8\textwidth}{!}{
\begin{tabular}{cccccc}
\hline
\textbf{Method}              & \textbf{\begin{tabular}[c]{@{}c@{}}Avg. \\ ACC\end{tabular}} & \textbf{\begin{tabular}[c]{@{}c@{}}Avg. \\ Specificity\end{tabular}} & \textbf{\begin{tabular}[c]{@{}c@{}}Avg. \\ Sensitivity\end{tabular}} & \textbf{\begin{tabular}[c]{@{}c@{}}Avg. \\ F1-score\end{tabular}} & \textbf{\begin{tabular}[c]{@{}c@{}}Avg. \\ Precision\end{tabular}} \\ \hline
        \textbf{KNN} & 0.5583 &  0.5498 & 0.5234 & 0.5100 & 0.4972 \\
        \textbf{Random Forest} & 0.6655 & 0.5803 & 0.5520 & 0.5302 & 0.5104\\
        \textbf{XGBoost} & 0.7501 & 0.7235 & 0.6820 & 0.6414 & 0.6054 \\ 
        \textbf{SVM} & 0.7901 & 0.7750 & 0.7200 & 0.7250 & 0.7300 \\ 
         \textbf{Syn-XRF} & 0.7018 & 0.8782 & 0.6143 & 0.6184 & 0.6225 \\
          \textbf{DNN} & 0.6650 & 0.5830 & 0.5620 & 0.5358 & 0.5120 \\
         \textbf{CBAM} & 0.9007 & 0.9023 & 0.9020 & 0.7848 & 0.6945\\ \hline
        \textbf{CAVE-Net} & \textbf{0.9108} & \textbf{0.9903} & \textbf{0.9108} & \textbf{0.7729} & \textbf{0.6928} \\ \hline
\end{tabular}}
\end{table*}

\begin{table*}[htbp]
\centering
\caption{Comparison of CAVE-Net Performance Metrics in the Validation Dataset.}
\label{tab:results2}
\resizebox{0.8\textwidth}{!}{
\begin{tabular}{cccccc}
\hline
\textbf{Method}              & \textbf{\begin{tabular}[c]{@{}c@{}}Avg. \\ ACC\end{tabular}} & \textbf{\begin{tabular}[c]{@{}c@{}}Avg. \\ Specificity\end{tabular}} & \textbf{\begin{tabular}[c]{@{}c@{}}Avg. \\ Sensitivity\end{tabular}} & \textbf{\begin{tabular}[c]{@{}c@{}}Avg. \\ F1-score\end{tabular}} & \textbf{\begin{tabular}[c]{@{}c@{}}Avg. \\ Precision\end{tabular}} \\ \hline
        \textbf{KNN} & 0.5634 &  0.8934 & 0.3839 & 0.2726 & 0.3831 \\
        \textbf{Random Forest} & 0.6352 & 0.9643 & 0.4944 & 0.4316 & 0.3830 \\
        \textbf{XGBoost} & 0.7459 & 0.9845 & 0.4036 & 0.3752 & 0.3506 \\ 
        \textbf{SVM} & 0.7882 & 0.9699 & 0.5340 & 0.4844 &  0.4432 \\ 
         \textbf{Syn-XRF} & 0.7089 & 0.9680 & 0.4330 & 0.3047 & 0.2351 \\
          \textbf{DNN}    & 0.6477 & 0.9975 & 0.5075 & 0.4588 & 0.4198 \\
         \textbf{CBAM}   & 0.8935 & 0.8924 & 0.8924 & 0.7614 & 0.6640\\ \hline
        \textbf{CAVE-Net} & \textbf{0.8993} & \textbf{0.9894} & \textbf{0.8993} & \textbf{0.7543} & \textbf{0.6722} \\
        \hline
\end{tabular}}
\end{table*}

Table \ref{tab:results3} depicts validation results of CAVE-Net in comparison to the baseline methods including VGG19, Xception, ResNet50V2, MobileNetV2, InceptionV3 and InceptionResNetV2, as reported by the organizing team of Capsule Vision 2024 challenge.

\begin{table*}[htbp]
\centering
\caption{Testing Results and Comparison to the Baseline Methods.}
\label{tab:results3}
\footnotesize
\begin{tabular}{cccc}
\hline
\textbf{Method}              & \textbf{\begin{tabular}[c]{@{}c@{}}Avg. \\ AUC\end{tabular}} & \textbf{\begin{tabular}[c]{@{}c@{}}Avg. \\ ACC\end{tabular}} & \textbf{\begin{tabular}[c]{@{}c@{}}Combined\\ Metric\end{tabular}} \\ \hline
\textbf{VGG19} & 0.5255   &     0.1445     & 0.3349\\
\textbf{Xception} & 0.5341 &   0.1313       & 0.3327\\
\textbf{ResNet50V2} & 0.5422    &   0.1773       & 0.3597\\
\textbf{MobileNetV2} & 0.5485    &    0.1140      & 0.3312\\
\textbf{InceptionV3} & 0.5250    &   0.1284       & 0.3267\\
\textbf{InceptionResNetV2}  & 0.5232    &     0.1469     & 0.33505\\ \hline
\textbf{CAVE-Net} & \textbf{0.7271} & \textbf{0.3359}  & \textbf{0.5315}\\ \hline
\end{tabular}
\end{table*}

The heatmaps visualize the confusion matrix, indicating the frequency of true labels (y-axis) against predicted labels (x-axis), as shown in Figure 3(a) and Figure 3(b), for training and validation data, respectively. 

\begin{figure}[H]
    \centering
    \caption{Heatmaps visualizing the confusion matrix for the dataset.}

    \begin{subfigure}[b]{0.45\textwidth} 
        \centering
        \includegraphics[width=\textwidth]{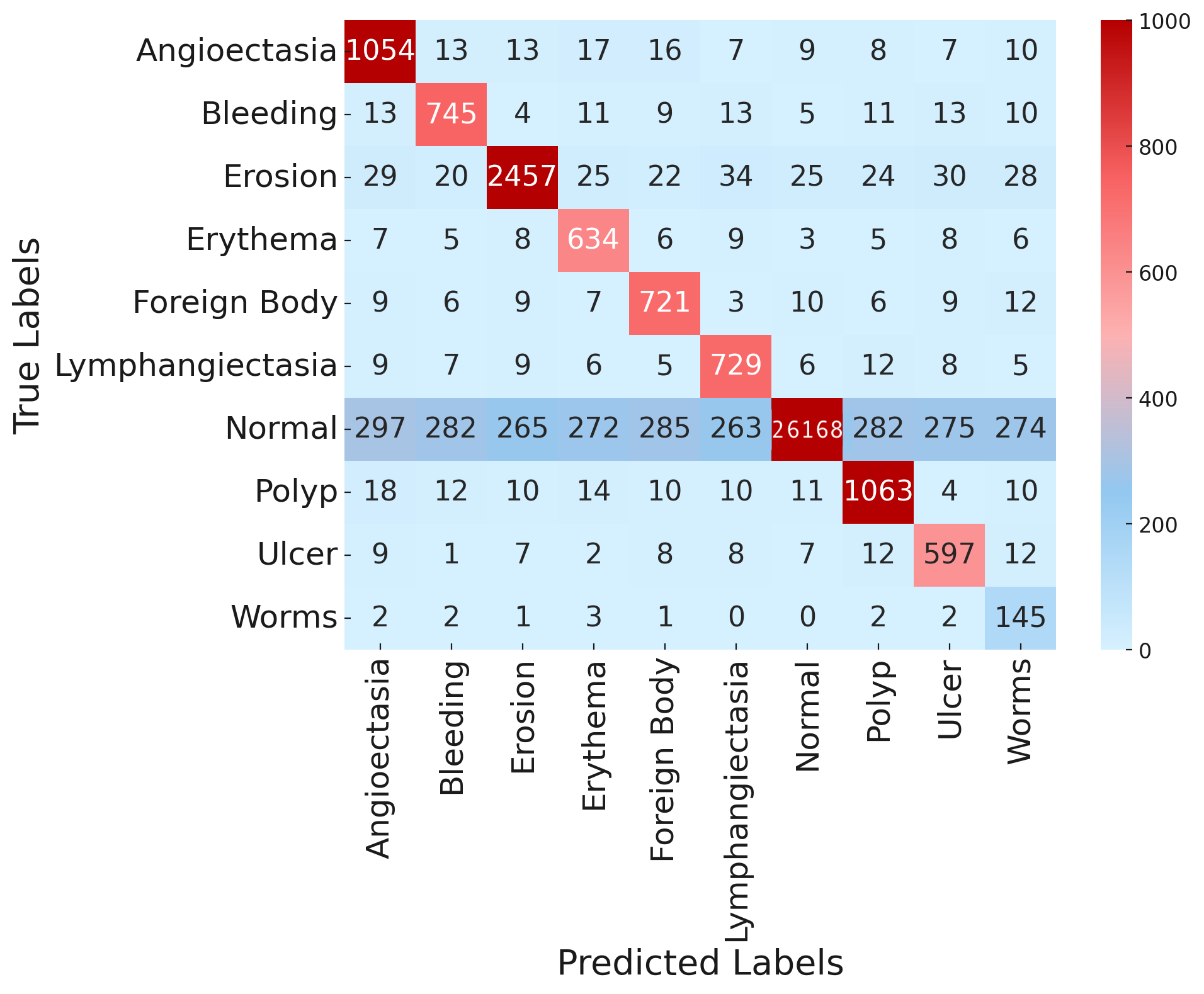} 
        \caption{Heatmap for training data.}
        \label{fig:sub1}
    \end{subfigure}
    \hspace{-0.03\textwidth} 
    \begin{subfigure}[b]{0.45\textwidth} 
        \centering
        \includegraphics[width=\textwidth]{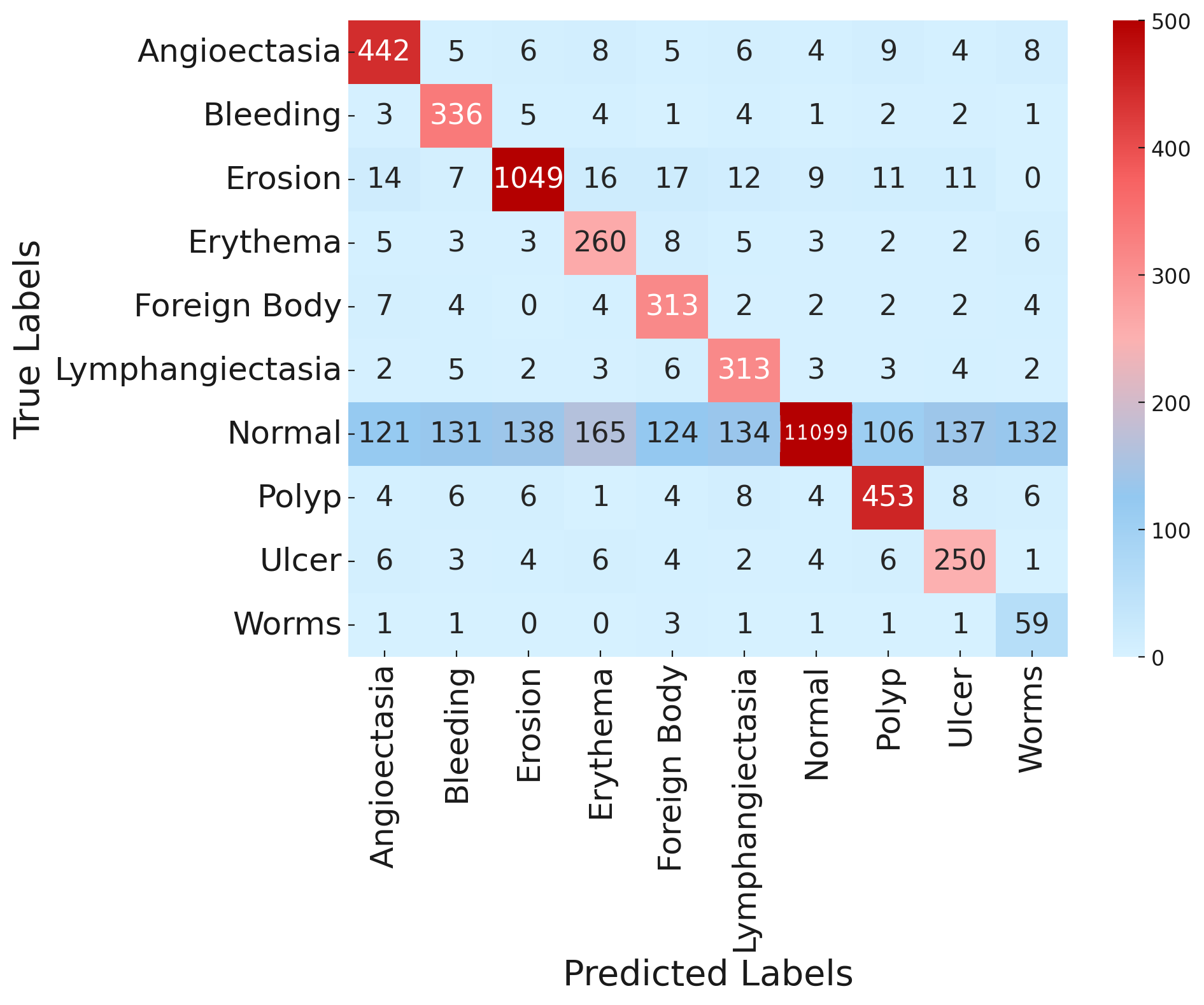} 
        \caption{Heatmap for validation data.}
        \label{fig:sub2}
    \end{subfigure}

    \label{fig:main}
\end{figure}

\subsection{Ranking and Recognition}  
CAVE-Net is ranked at the $6^{th}$ position out of more than $150$ teams that participated globally in the Capsule Vision 2024 Challenge, with a mean AUC score of $0.7271482$ and a balanced accuracy score of $0.3359341$ over the test dataset as mentioned in~\cite{handa2024capsulevision2024challenge}.

\section{Discussion}\label{sec4}

Our study demonstrates that the ensemble approach integrating CBAM, DNN and traditional machine learning models in Syn-XRF, significantly enhances classification performance in multiclass abnormality detection, particularly in medical imaging.
The CBAM module effectively highlights critical features, improving the model’s ability to detect subtle but significant differences of specific conditions.
The DNN architecture complements this by learning complex patterns, while ensemble-based traditional models approach, Syn-XRF, manage issues such as over fitting and class imbalance.
Their combination for the final classification increases the accuracy.
The Syn-XRF is more appropriate for handling high-dimensional and variable-quality data, representing a significant advancement in automated diagnosis.
This approach has the potential to greatly enhance medical diagnostics by reducing diagnostic errors and expediting patient care.
Continued refinement of this model can lead to even greater precision in automated diagnosis and significantly benefit the medical field.

\section{Conclusion}\label{sec5}

We have used a combination of Convolutional Block Attention Module, Deep Neural Network, and ensemble techniques such as RF, XGBoost, SVM and KNN at Syn-XRF.
This combination effectively leverages the strengths of each individual model, resulting in improved accuracy and robustness in classification tasks.
The synergy created by the proposed framework, CAVE-Net, ensemble approach demonstrates its superiority over the other frameworks examined, underscoring its potential for practical applications in automated diagnosis.

\section{Acknowledgments}\label{sec6}
As participants in the Capsule Vision 2024 Challenge, we fully comply with the competition's rules as outlined in \cite{handa2024capsulevision2024challenge}. Our AI model development is based exclusively on the datasets provided in the official release in \cite{Handa2024Training,Handa2024}.
Additionally, we would like to extend our gratitude to the Centre of Excellence (CoE) in Artificial Intelligence (AI) of Netaji Subhas University of Technology (NSUT), New Delhi, for providing the resources and support needed for this research.

\bibliographystyle{unsrtnat}
\bibliography{main}

\end{document}